\documentclass{article}

\usepackage{microtype}
\usepackage{graphicx}
\usepackage{booktabs}
\usepackage{array}
\usepackage{hyperref}
\usepackage[preprint]{icml2026}
\usepackage{amsmath}
\usepackage{amssymb}
\usepackage{mathtools}
\usepackage{amsthm}
\usepackage[capitalize,noabbrev]{cleveref}

\theoremstyle{definition}
\newtheorem{definition}{Definition}[section]

\icmltitlerunning{Answer Only as Precisely as Justified}

\newcommand{\css}{\textnormal{\textsc{CSS}}}
\newcommand{\oau}{\textnormal{\textsc{OAU}}}
\newcommand{\fine}{\mathrm{fine}}
\newcommand{\coarse}{\mathrm{coarse}}
\newcommand{\omitc}{\mathrm{omit}}
\newcommand{\cpu}{\mathrm{CPUpper}}
\newcommand{\ind}{\mathbf{1}}
\newcolumntype{L}[1]{>{\raggedright\arraybackslash}p{#1}}

\begin{document}

\twocolumn[
\icmltitle{Answer Only as Precisely as Justified: Calibrated Claim-Level Specificity Control for Agentic Systems}

\begin{icmlauthorlist}
\icmlauthor{Tianyi Huang}{ryq}
\icmlauthor{Samuel Xu}{appin}
\icmlauthor{Jason Tansong Dang}{appin}
\icmlauthor{Samuel Yan}{appin}
\icmlauthor{Kimberley Yin}{appin}
\end{icmlauthorlist}

\icmlaffiliation{ryq}{Ryquo}
\icmlaffiliation{appin}{App-In Club}

\icmlcorrespondingauthor{Tianyi Huang}{tianyi@ryquo.com}
\icmlkeywords{agentic systems, uncertainty calibration, factuality, claim-level verification, reliability}

\vskip 0.3in
]

\printAffiliationsAndNotice{}

\begin{abstract}
Agentic systems often fail not by being entirely wrong, but by being too precise: a response may be generally useful while particular claims exceed what the evidence supports. We study this failure mode as \emph{overcommitment control} and introduce \emph{compositional selective specificity} (\css), a post-generation layer that decomposes an answer into claims, proposes coarser backoffs, and emits each claim at the most specific calibrated level that appears admissible. The method is designed to express uncertainty as a local semantic backoff rather than as a whole-answer refusal. Across a full LongFact run and HotpotQA pilots, calibrated \css\ improves the risk-utility trade-off of fixed drafts. On the full LongFact run, it raises overcommitment-aware utility from 0.846 to 0.913 relative to the no-\css\ output while achieving 0.938 specificity retention. These results suggest that claim-level specificity control is a useful uncertainty interface for agentic systems and a target for future distribution-free validity layers.
\end{abstract}

\section{Introduction}

Modern language-model systems increasingly retrieve evidence, call tools, and assemble multi-claim artifacts for users or downstream components \citep{lewis2020rag,karpas2022mrkl,yao2023react,schick2023toolformer}. In such systems, uncertainty is not simply a confidence score attached to a final string. It is a control signal that should influence what the system says, how specifically it says it, and which semantic commitments are safe to propagate. This motivates a shift from answer-level uncertainty to \emph{compositional uncertainty}: uncertainty over the claims that make up an agent output.

A common failure mode in this setting is \emph{overcommitment}. The response may be directionally correct but unjustifiably specific: it may identify the relevant event while attaching an unsupported date, summarize the right trend while supplying an exact number the evidence does not warrant, or name the correct entity while asserting an unsupported subtype or relation. Outright refusal is often too conservative in these cases, yet the original wording still overstates what is known. The practical question is therefore not only whether the model should answer, but also how much semantic precision it should retain for each emitted claim.

This question sits at the intersection of selective prediction, calibration, and factuality control. Selective prediction and conformal prediction provide principled ways to abstain or broaden prediction sets when uncertainty is high \citep{geifman2017selective,angelopoulos2023conformal,quach2024conformal,mohri2024conformal}. Long-form factuality work has shown that answer-level judgments are often too coarse because generations mix supported and unsupported statements within the same response \citep{min2023factscore,wei2024longfact}. Recent specificity-aware methods further suggest that less-specific rewrites can be more useful than all-or-nothing abstention \citep{jiang2025clc,goren2026selective}. We build on this line of work, but focus on agentic outputs as compositional objects whose parts may warrant different levels of commitment.

We propose \emph{compositional selective specificity} (\css). Given a draft answer, \css\ decomposes it into claims. For each claim, it constructs a small specificity ladder consisting of the original fine-grained statement, a coarser backoff, and omission. A support estimator scores which candidates appear justified by the available evidence, and a selector emits the most specific admissible option. The resulting response can remain precise where evidence is strong, back off where support is partial, and omit only claims whose specificity cannot be justified. Uncertainty is thereby expressed as a structured editing policy over semantic granularity rather than as vague hedging or global refusal.

The deployed method in this paper is \emph{calibrated \css}. Calibration is the operational core: it turns noisy support scores into a claim-selection rule. We compare calibrated \css\ against no post-processing, whole-answer abstention, claim deletion, and a fixed-threshold \css\ ablation. We also report oracle \css\ as a non-deployable upper bound, using it to estimate the headroom available if support estimation were nearly perfect.

Our empirical evidence comes from two benchmarks: LongFact and HotpotQA \cite{wei2024longfact, yang2018hotpotqa}. The complete LongFact run is the main scale evaluation; it compares calibrated \css\ against the above baselines on all 2,280 LongFact prompts under a claimwise protocol. The LongFact pilot and the GPT-5.4 HotpotQA pilot provide smaller controlled checks of the calibration effect on fixed drafts. The Claude Sonnet 4.6 HotpotQA pilot provides an auxiliary check under a different frontier model.

The paper makes three main contributions. First, it formulates overcommitment control as a claim-level uncertainty problem for agentic outputs. Second, it presents a black-box \css\ pipeline whose deployable form is a calibrated selector over claim-level backoff choices. Third, it provides evidence that calibration can preserve substantially more supported specificity than fixed-threshold or coarse-abstention alternatives, especially in the long-form multi-claim settings. We view these results as a step toward, and not a substitute for, stronger finite-sample guaranties.

\section{Related Work}

\paragraph{Selective prediction and conformal uncertainty.}
Selective prediction equips a model with a reject option, trading coverage for lower risk on retained examples \citep{geifman2017selective}. Conformal prediction generalizes this idea by wrapping black-box predictors with distribution-free uncertainty sets and abstention mechanisms \citep{angelopoulos2023conformal}. For language generation, conformal language modeling adapts this viewpoint to structured natural-language outputs, while conformal factuality studies correctness guarantees for language-model responses \citep{quach2024conformal,mohri2024conformal}. Our setting is adjacent but distinct: rather than accepting or rejecting an entire response, we control the granularity of each claim within that response.

\paragraph{Long-form factuality and verification.}
Long-form generations routinely contain mixtures of supported and unsupported content, which makes binary answer-level judgments inadequate. FActScore formalizes factual precision through atomic facts, and LongFact scales this perspective with claim decomposition and search-based checking \citep{min2023factscore,wei2024longfact}. Verification-oriented methods such as SelfCheckGPT and Chain-of-Verification similarly emphasize post-hoc verification rather than raw generation confidence \citep{manakul2023selfcheckgpt,dhuliawala2024cove}. Our work shares the claimwise lens, but asks a different control question: once support is heterogeneous within a response, how should the answer be edited so that unsupported specificity is removed while useful content is preserved?

\paragraph{Specificity-aware control.}
The closest prior work studies uncertainty-aware rewrites that trade specificity for reliability. Conformal Linguistic Calibration connects factuality and imprecision through a conformal answer-set view, while Selective Abstraction shows that less-specific reformulations can outperform pure claim removal in long-form generation \citep{jiang2025clc,goren2026selective}. We are sympathetic to both views. Our emphasis differs in two ways. First, we target agentic deployment: the output is a structured allocation of semantic commitment that downstream modules could use for escalation, auditing, or re-retrieval. Second, we emphasize calibration as the deployable mechanism that selects a useful operating point for a fixed support estimator.

\section{Problem Setup}

Let $x$ denote a prompt, $E$ the evidence available for that prompt, and $A$ a draft answer produced by an upstream generator. A claim extractor maps $A$ to atomic claims,
\begin{equation}
A \mapsto (c_1,\ldots,c_m).
\end{equation}
In our implementation, each claim $c_i$ is paired with a coarse backoff $\tilde{c}_i$ and omission as the null action. Thus each claim has three possible output levels,
\begin{equation}
\pi_i \in \{\fine,\coarse,\omitc\}.
\end{equation}
The fine level emits the original claim, the coarse level emits a less-specific reformulation, and the omit level emits nothing.

A support estimator assigns scores $s_i^{\fine},s_i^{\coarse}\in[0,1]$ to the non-null candidates. For offline evaluation, we also obtain binary support labels $y_i^{\fine},y_i^{\coarse}\in\{0,1\}$ indicating whether the corresponding formulation is supported by the evidence. These labels are used only for scoring and for the oracle upper bound; they are not available to the deployable selector.

\begin{definition}[Overcommitment]
A selected claim is \emph{overcommitted} if it is emitted at the chosen specificity level but unsupported by the evidence at that level.
\end{definition}

The no-\css\ identity policy keeps every original fine-grained claim. \css\ instead chooses one action $\pi_i$ for each claim. We define specificity weights
\begin{equation}
w(\omitc)=0,\qquad w(\coarse)=\gamma,\qquad w(\fine)=1,
\end{equation}
with $\gamma=0.6$ in all experiments. For convenience, define $y_i^{\omitc}=0$, let $e_i=\ind\{\pi_i\neq\omitc\}$, and write $y_i^{\mathrm{sel}}:=y_i^{\pi_i}$ for the support label of the emitted level. Aggregating over all claims in an evaluation split, we report
\begin{align}
\mathrm{Prec} &= \frac{\sum_i e_i y_i^{\mathrm{sel}}}{\sum_i e_i}, \label{eq:precision}\\
\mathrm{Ret} &= \frac{1}{m}\sum_i w(\pi_i), \label{eq:retention}\\
\mathrm{SuppSpec} &= \frac{1}{m}\sum_i w(\pi_i)y_i^{\mathrm{sel}}, \label{eq:suppspec}\\
\oau &= \frac{1}{m}\sum_i \left(w(\pi_i)y_i^{\mathrm{sel}}-e_i(1-y_i^{\mathrm{sel}})\right). \label{eq:oau}
\end{align}
Precision is defined when at least one claim is emitted, as in all reported runs. The \oau\ score rewards supported specificity, assigns zero to omitted claims, and penalizes unsupported emitted claims.

\section{Method: Calibrated Compositional Selective Specificity}

\subsection{Pipeline}

\css\ is a post-generation layer. It does not retrain the upstream model or alter the retrieval stack. Instead, it edits a fixed draft answer through four stages.

\paragraph{Draft generation.}
An upstream language model produces an initial response given the prompt and available evidence.

\paragraph{Claim extraction and backoff proposal.}
The draft answer is decomposed into individual claims. For each claim, the system proposes a coarse backoff that preserves the central content while removing potentially unsupported detail. The selector then chooses among three actions: keep the fine claim, emit the coarse backoff, or omit the claim.

\paragraph{Support scoring.}
Each fine or coarse candidate is evaluated against the evidence. Our implementation combines language-model-based support judgments with lightweight lexical and entity-based features through a hybrid scorer. Its precise combiner is not the contribution here: within each run, the scorer is fit once and then held fixed across all policy comparisons. The contribution of this paper is the selection layer that consumes those support scores, not a new verifier.

\paragraph{Claimwise selection.}
The selector chooses the level for each claim. If the fine claim clears the fine threshold, it is kept. Otherwise, if the coarse backoff clears the coarse threshold, the claim is emitted in coarse form. If neither clears, the claim is omitted.

\begin{figure*}[t]
\centering
\includegraphics[width=0.95\textwidth]{example.png}
\caption{Illustrative behavior of claim-level specificity control. The evidence supports that the agreement was signed in Geneva, but does not support the exact year in the draft. A whole-answer abstention policy returns no answer, while a conservative fixed-threshold \css\ selector may omit the claim. Calibrated \css\ can instead back off only the unsupported detail and preserve the supported claim at a coarser specificity level. The example illustrates the intended role of \css: expressing uncertainty through localized semantic backoff rather than global refusal or unsupported precision.}
\label{fig:css_method_example}
\end{figure*}

\subsection{Policies and Baselines}

\paragraph{No \css.}
This identity baseline emits every original fine-grained claim. It measures how much overcommitment is present before claim-level control.

\paragraph{Whole-answer abstention.}
This coarse baseline applies a response-level accept-or-abstain decision. Operationally, it either returns the draft unchanged or emits no answer at all; it does not perform claimwise editing. It is included to quantify the cost of treating uncertainty as a whole-answer event rather than a claimwise editing problem.

\paragraph{Claim-drop.}
This baseline omits claims that do not clear a support threshold but does not emit a coarser alternative. Operationally, it either keeps the original fine-grained claim or deletes it; it never emits a coarse backoff. It tests whether \css\ helps beyond deletion by preserving partially supported content through backoff.

\paragraph{Uncalibrated \css.}
This deployable ablation uses fixed thresholds for all runs. In our implementation, the defaults are $\tau_{\fine}=0.78$ and $\tau_{\coarse}=0.58$. It does not adapt thresholds to the score distribution of a run.

\paragraph{Calibrated \css.}
This is the deployed method. A held-out calibration split is used to choose thresholds $(\tau_{\fine},\tau_{\coarse})$ that maximize utility subject to a conservative upper bound on unsupported emissions. For each threshold pair, we count unsupported emitted claims $k$ and emitted claims $n$ on the calibration set, form a one-sided Clopper--Pearson upper confidence bound \citep{clopper1934confidence}, and keep only pairs satisfying
\begin{equation}
\cpu(k,n;\delta) \leq \alpha,
\end{equation}
with $\alpha=0.10$ and $\delta=0.05$ in the reported runs. In the reported runs, at least one threshold pair satisfied this constraint, so the degenerate no-valid-pair case did not arise. Among valid pairs, the selector chooses the one with the highest calibration \oau. This makes calibration explicit, conservative, and aligned with the claim-level utility used in evaluation. Because the same calibration split is used to select among threshold pairs, we treat this step as a conservative calibration rule rather than as a full end-to-end conformal guarantee.

\paragraph{Oracle \css.}
This non-deployable upper bound selects the fine claim if it is evaluation-supported, otherwise the coarse claim if the coarse version is evaluation-supported, otherwise omission. Oracle \css\ estimates the headroom available if support estimation were much stronger; it is not a competing method.

Two points are important for interpretation. First, calibrated \css\ is the method we advocate as a deployable uncertainty layer. Second, all policies operate on fixed drafts within a run. The relevant comparison is therefore between post-generation selection policies, not between different generators.

\section{Experimental Setup}

\paragraph{LongFact full run.}
Our main experiment uses all 2,280 LongFact prompts introduced by \citet{wei2024longfact}. Because our interest is claim-level specificity control rather than the SAFE/F1@K benchmark pipeline, we evaluated these prompts with the claimwise protocol. The run uses a five-fold out-of-fold evaluation: every prompt is scored in a held-out fold, while scorer fitting and threshold calibration are performed on the remaining folds. The completed run contains 11,705 extracted claims and uses GPT-5.4, accessed through the OpenAI API, as the draft generator \citep{openai2026gpt54}.

\paragraph{Pilot runs.}
We also report a 200-example LongFact pilot and two 200-example HotpotQA validation pilots \citep{yang2018hotpotqa}. Each pilot is split into 30 fit, 30 calibration, and 140 test examples. The LongFact pilot contains 757 extracted claims on the test split; the GPT-5.4 HotpotQA pilot contains 470 claims \citep{openai2026gpt54}; and the Claude Sonnet 4.6 HotpotQA pilot, run through Anthropic's API, contains 648 claims \citep{anthropic2026sonnet46}. Claim counts differ because claim extraction is performed on generated drafts.

\paragraph{What is being compared?}
The full-run comparison asks whether calibrated \css\ improves the specificity--risk trade-off relative to no post-processing, whole-answer abstention, claim deletion, and fixed-threshold \css. The pilot comparisons ask whether the calibration effect replicates on smaller fixed-draft subsets. Oracle rows are reported as ceilings.

\paragraph{LongFact protocol.}
LongFact was introduced together with SAFE and F1@K as its official benchmark machinery \citep{wei2024longfact}. Our goal here is narrower: we evaluate claim-level post-generation editing on LongFact prompts. Concretely, extracted fine/coarse claim candidates are assigned binary support labels with respect to the prompt evidence, and all reported LongFact metrics are computed from these claim-level support labels. The LongFact numbers in this paper should therefore be read as results under the claimwise protocol, not as SAFE/F1@K leaderboard numbers.

\section{Results}

We begin with the full LongFact run, which is the paper's main scale evaluation. The no-\css\ draft keeps all 11,705 extracted claims and therefore achieves perfect specificity retention by construction, but only 0.9230 support precision and 0.8460 \oau. Calibrated \css\ instead emits 11,085 claims, raises support precision to 0.9865, and improves \oau\ to 0.9130 while still achieving 0.9381 specificity retention. The interesting point is that the selector becomes safer while preserving most of the useful content in the original drafts.

\begin{figure}[t]
\centering
\includegraphics[width=\columnwidth]{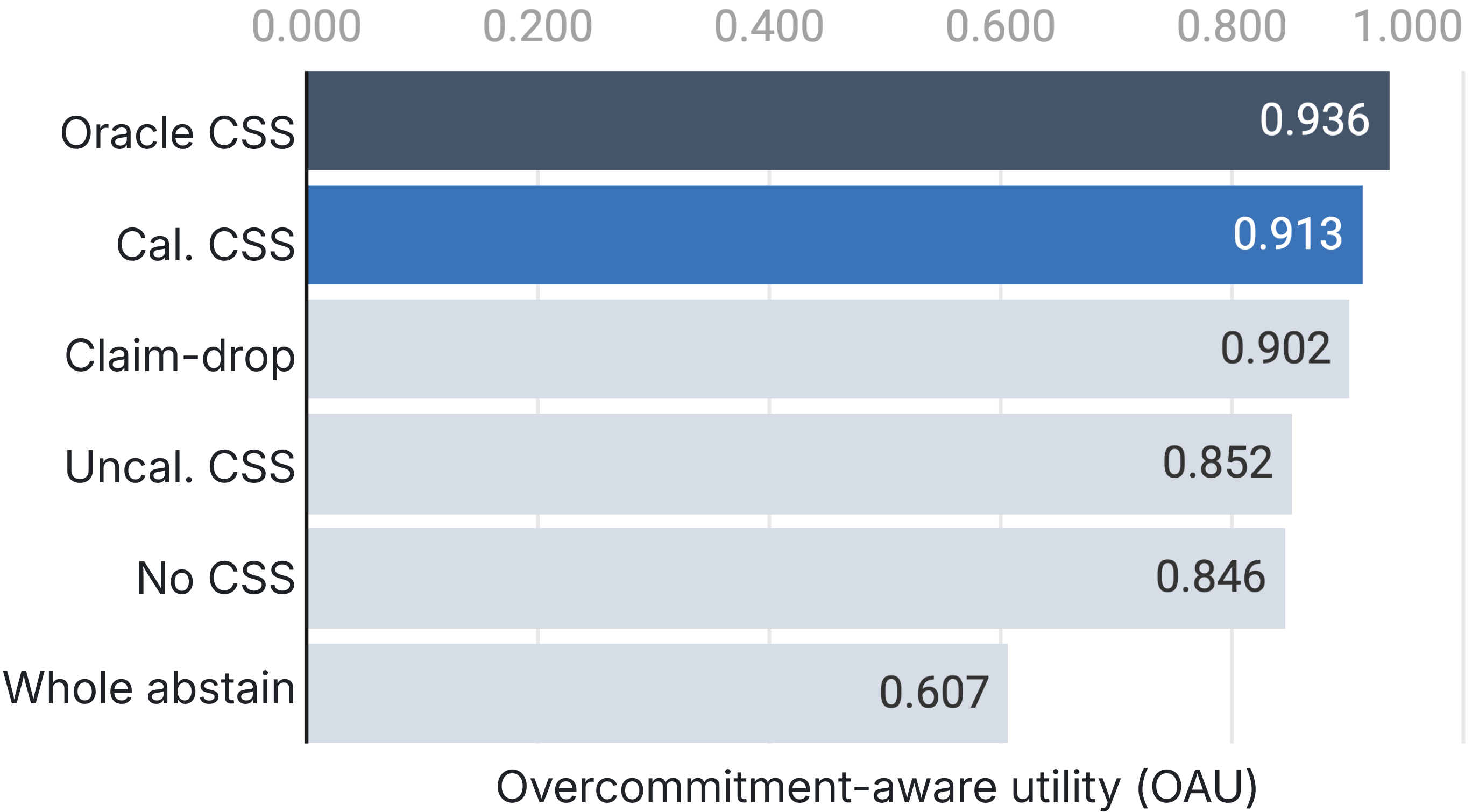}
\caption{Full LongFact policy comparison by overcommitment-aware utility (\oau). Calibrated \css\ is the deployed selector, gray bars are reference baselines, and the dark bar is the non-deployable oracle ceiling.}
\label{fig:longfact_oau}
\end{figure}

The baseline comparisons clarify what calibrated \css\ is buying. As \Cref{fig:longfact_oau} and \Cref{tab:main_results} show, whole-answer abstention is not a competitive route to reliability: although it attains high support precision, it discards too much content, reducing specificity retention to 0.6292 and \oau\ to 0.6072. Claim-drop is much stronger, reaching 0.9877 support precision and 0.9019 \oau, but calibrated \css\ still improves upon it by preserving partially supported content through coarse backoff rather than deleting it outright. Uncalibrated \css\ is the strongest conservative deployable baseline, with 0.9934 support precision, but it achieves only 0.8633 specificity retention and reaches 0.8521 \oau. Relative to this fixed-threshold selector, calibration gives up 0.0069 support-precision points while gaining 0.0609 \oau\ points and 0.0675 supported-specificity points.

This is the central empirical result of the paper. On the full LongFact run, a calibrated claim-level selector improves over the original draft, a whole-answer abstention baseline, a deletion-only baseline, and a fixed-threshold \css\ ablation on the same fixed full-run drafts. The relevant operating regime is therefore not perfect precision at any cost, but high precision together with high retained supported specificity.

\begin{table*}[t]
\caption{Claim-level results from the completed runs. Calibrated \css\ is the deployed method. Uncalibrated \css\ is the fixed-threshold ablation. Oracle \css\ is a non-deployable upper bound.}
\label{tab:main_results}
\centering
\small
\setlength{\tabcolsep}{3.6pt}
\begin{tabular*}{\textwidth}{@{\extracolsep{\fill}}L{2.55cm} L{1.75cm} c c c c c c@{}}
\toprule
Dataset / run & Policy & Ex. & \shortstack{Claims emitted\\ / total} & \shortstack{Support\\precision} & \shortstack{Specificity\\retention} & \shortstack{Supported\\specificity} & \oau\ \\
\midrule
LongFact full / GPT-5.4 & No \css & 2280 & 11705 / 11705 & 0.9230 & 1.0000 & 0.9230 & 0.8460 \\
LongFact full / GPT-5.4 & Whole abstain & 2280 & 7365 / 11705 & 0.9825 & 0.6292 & 0.6182 & 0.6072 \\
LongFact full / GPT-5.4 & Claim-drop & 2280 & 10823 / 11705 & 0.9877 & 0.9246 & 0.9133 & 0.9019 \\
LongFact full / GPT-5.4 & Uncalibrated & 2280 & 10948 / 11705 & 0.9934 & 0.8633 & 0.8583 & 0.8521 \\
LongFact full / GPT-5.4 & Calibrated & 2280 & 11085 / 11705 & 0.9865 & 0.9381 & 0.9258 & 0.9130 \\
LongFact full / GPT-5.4 & Oracle & 2280 & 11056 / 11705 & 1.0000 & 0.9359 & 0.9359 & 0.9359 \\
\midrule
LongFact pilot / GPT-5.4 & Uncalibrated & 140 & 721 / 757 & 0.9958 & 0.5900 & 0.5876 & 0.5836 \\
LongFact pilot / GPT-5.4 & Calibrated & 140 & 724 / 757 & 0.9931 & 0.9411 & 0.9355 & 0.9289 \\
LongFact pilot / GPT-5.4 & Oracle & 140 & 723 / 757 & 1.0000 & 0.9440 & 0.9440 & 0.9440 \\
HotpotQA pilot / GPT-5.4 & Uncalibrated & 140 & 409 / 470 & 0.9829 & 0.6209 & 0.6111 & 0.5962 \\
HotpotQA pilot / GPT-5.4 & Calibrated & 140 & 429 / 470 & 0.9790 & 0.9000 & 0.8809 & 0.8617 \\
HotpotQA pilot / GPT-5.4 & Oracle & 140 & 426 / 470 & 1.0000 & 0.8996 & 0.8996 & 0.8996 \\
HotpotQA pilot / Claude Sonnet 4.6 & Uncalibrated & 140 & 622 / 648 & 0.9936 & 0.7191 & 0.7142 & 0.7080 \\
HotpotQA pilot / Claude Sonnet 4.6 & Calibrated & 140 & 628 / 648 & 0.9904 & 0.9475 & 0.9395 & 0.9302 \\
HotpotQA pilot / Claude Sonnet 4.6 & Oracle & 140 & 630 / 648 & 1.0000 & 0.9599 & 0.9599 & 0.9599 \\
\bottomrule
\end{tabular*}
\end{table*}

\subsection{Calibration on Pilots}

The pilot runs ask a narrower question: given fixed drafts, does calibration materially improve the selector? Across all three pilots, the answer is yes.

On the LongFact pilot, calibration raises specificity retention from 0.5900 to 0.9411 and \oau\ from 0.5836 to 0.9289, while support precision decreases only from 0.9958 to 0.9931. On the GPT-5.4 HotpotQA pilot, calibration raises specificity retention from 0.6209 to 0.9000, supported specificity from 0.6111 to 0.8809, and \oau\ from 0.5962 to 0.8617, while support precision moves only from 0.9829 to 0.9790. The Claude Sonnet 4.6 HotpotQA pilot shows a similar pattern: calibration increases specificity retention from 0.7191 to 0.9475, supported specificity from 0.7142 to 0.9395, and \oau\ from 0.7080 to 0.9302, while support precision decreases from 0.9936 to 0.9904.

The consistency of this pattern matters more than any single pilot in isolation. In all three cases, the fixed-threshold selector identifies an ultra-safe but overly conservative operating point. Calibration then moves the selector to a much more useful part of the trade-off curve, preserving substantially more supported detail while paying only a small precision cost. This is a lesson of the paper: calibration is not a cosmetic refinement, but the mechanism that converts noisy claimwise support scores into a practical uncertainty layer.

\subsection{Oracle Ceilings Quantify Headroom}

Oracle \css\ should be interpreted only as a ceiling. On the full LongFact run, oracle \css\ reaches 0.9359 \oau\ with perfect support precision, compared with 0.9130 for calibrated \css. This gap is meaningful but not large, suggesting that the calibrated selector already captures a substantial fraction of the available benefit and that further gains will depend primarily on stronger support estimation. The pilot ceilings tell a similar story: oracle \css\ reaches 0.9440 \oau\ on the LongFact pilot, 0.8996 on the GPT-5.4 HotpotQA pilot, and 0.9599 on the Claude Sonnet 4.6 HotpotQA pilot.

These results show that the underlying claim-level backoff problem is structurally favorable: when support information is sufficiently accurate, a selector can preserve most of a response's specificity without paying a penalty in unsupported detail.

\subsection{What the Completed Runs Show}

Taken together, the completed runs support four conclusions. First, claim-level specificity control remains effective at the full LongFact scale. Second, calibration is the central deployable lever: across the full run and all three pilots, calibrated \css\ consistently trades a small amount of precision for a much larger gain in retained supported specificity. Third, coarse answer-level abstention and deletion-only control are informative baselines, but they are not adequate substitutes for compositional specificity control. Fourth, the oracle ceilings indicate that meaningful headroom remains, especially if support scoring improves. 

\section{Discussion and Limitations}

The paper's framing has two main implications. Methodologically, it suggests that answer-level accuracy is often the wrong object for evaluating post-generation uncertainty layers. The more relevant question is whether the system says \emph{only as much as it knows}. Our metrics make that question explicit by separating retained specificity from supported specificity and by penalizing unsupported emitted claims.

Systems-wise, claim-level specificity control creates an uncertainty interface that downstream components can use directly. Claims that survive at fine granularity can move forward with less friction; claims that are downgraded or omitted can trigger re-retrieval, human review, or escalation to a stronger verifier. In this sense, \css\ is a modular control layer for multi-stage systems.

Several limitations remain. First, the LongFact experiment uses the claimwise evaluation pipeline on LongFact prompts rather than the SAFE/F1@K pipeline \citep{wei2024longfact}. This is appropriate for the paper's goal---evaluating claim-level post-generation editing rather than whole-response benchmark ranking---but it means the LongFact numbers should be interpreted accordingly. Second, the pilot comparisons use 200-example subsets with 140 evaluated test examples. Third, oracle \css\ is only an upper bound and should not be mistaken for deployable performance. Fourth, the selector can inherit errors from claim extraction, backoff generation, and support estimation. Finally, the present paper does not study sequential or continuously monitored settings.

We view these limitations as next steps in the problem formulation. The completed runs suggest that claimwise specificity control is a meaningful and measurable uncertainty problem. The next stage is to attach stronger statistical guarantees to this control layer and to evaluate it inside broader agent pipelines with tool use, escalation, and external feedback loops.

\section{Conclusion}

Many agentic failures are failures of specificity, not simply failures of correctness. This paper treats that observation as a control problem: for each claim in a generated response, decide the most precise formulation justified by the available evidence. Calibrated \css\ is a step in this direction. It is empirically useful on the completed LongFact and HotpotQA runs, while still leaving room for stronger support estimation and formal risk guarantees. We hope the framing helps connect long-form factuality, selective prediction, and uncertainty-aware agent design around a common question: not only whether a system should answer, but how precise the truth allows it to be.

\section*{Impact Statement}

This work studies a mechanism for reducing unsupported specificity in language-model outputs. Such mechanisms can improve reliability in reporting, question answering, and decision support by preventing systems from stating more than their evidence warrants. However, claim-level backoff could also be misused to make weak systems appear trustworthy through vagueness alone. For that reason, we emphasize joint reporting of support precision, retained specificity, and supported specificity, rather than rewarding imprecision for its own sake. We do not view \css\ as a substitute for domain validation, expert review, or stronger verification in high-stakes settings.

\bibliographystyle{icml2026}
\bibliography{main}

\end{document}